\documentclass{article}

\usepackage{microtype}
\usepackage{graphicx}
\usepackage{subfigure}
\usepackage{booktabs} 

\usepackage{hyperref}

\usepackage[accepted]{icml2024}

\usepackage{amsmath}
\usepackage{amssymb}
\usepackage{mathtools}
\usepackage{amsthm}

\usepackage{comment}
\usepackage{color,soul}

\usepackage{multirow}
\usepackage{tabularx}
\usepackage{caption}
\usepackage{array}

\usepackage{graphicx}

\usepackage[capitalize,noabbrev]{cleveref}
\usepackage{listings} 

\theoremstyle{plain}

\theoremstyle{definition}

\theoremstyle{remark}

\usepackage[textsize=tiny]{todonotes}

\icmltitlerunning{Deep Content Understanding Toward Entity and Aspect Target Sentiment Analysis on Foundation Models}

\begin{document}

\lstset{
  basicstyle=\ttfamily, 
  breaklines=true, 
  frame=single, 
  columns=fullflexible 
}

\twocolumn[
\icmltitle{Deep Content Understanding Toward Entity and Aspect Target Sentiment Analysis on Foundation Models}



\icmlsetsymbol{equal}{*}

\begin{icmlauthorlist}
\icmlauthor{Vorakit Vorakitphan}{ibmc}
\icmlauthor{Milos Basic}{ibmc}
\icmlauthor{Guilhaume Leroy Meline}{ibmc}

\end{icmlauthorlist}

\icmlaffiliation{ibmc}{IBM Consulting France}
\icmlcorrespondingauthor{Vorakit Vorakitphan}{vorakit.vorakitphan1@ibm.com}

\icmlkeywords{Sentiment Analysis, Aspect-based Sentiment Analysis, Generative AI}

\vskip 0.3in
]



\printAffiliationsAndNotice{}  

\begin{abstract}
Introducing Entity-Aspect Sentiment Triplet Extraction (EASTE), a novel Aspect-Based Sentiment Analysis (ABSA) task which extends Target-Aspect-Sentiment Detection (TASD) by separating aspect categories (e.g., food\#quality) into pre-defined entities (e.g., meal, drink) and aspects (e.g., taste, freshness) which add a fine-gainer level of complexity, yet help exposing true sentiment of chained aspect to its entity. We explore the task of EASTE solving capabilities of language models based on transformers architecture from our proposed unified-loss approach via token classification task using BERT architecture to text generative models such as Flan-T5, Flan-Ul2 to Llama2, Llama3 and Mixtral employing different alignment techniques such as zero/few-shot learning, Parameter Efficient Fine Tuning (PEFT) such as Low-Rank Adaptation (LoRA). The model performances are evaluated on the SamEval-2016 benchmark dataset representing the fair comparison to existing works. Our research not only aims to achieve high performance on the EASTE task but also investigates the impact of model size, type, and adaptation techniques on task performance. Ultimately, we provide detailed insights and achieving state-of-the-art results in complex sentiment analysis.
\end{abstract}

\section{Introduction}
\label{sec:intro}
Sentiment Analysis (SA) is the field of Natural Language Processing (NLP) that aims to extract and analyze sentiments, opinions, attitudes and emotions expressed towards certain entities from a textual data. Traditionally, SA focused on detecting the overall polarity e.g., positive, negative or neutral. As such, it brings limited amount of information, often insufficient for most of the real world applications \cite{Liu_2012}. Hence, in the past decade, the research recognized fine-grained SA frequently named as Aspect Based Sentiment Analysis (ABSA) \cite{pontiki-etal-2014-semeval, pontiki-etal-2015-semeval, pontiki-etal-2016-semeval} whose task fundamentally consists of detecting two components which are targets and its corresponding sentiments.
According to \citet{zhang2022survey} the target can be described as either \textit{aspect category} or \textit{aspect term}, sentiment can be described as \textit{sentiment polarity} that often attaches by \textit{opinion term}. Depending on the motives, not all authors aim to extract these elements simultaneously, but they would rather focus on identifying specific subgroups and the relations among its elements which are consequently leading to the creation of various so called ABSA subtasks. 

However, the transformer architectures \cite{DBLP:journals/corr/VaswaniSPUJGKP17} with pre-trained knowledge (aka Foundation Models (FMs)) influence most of the NLP tasks to not require a specific architecture to train from scratch on a large corpus of data, but rather the best performing solutions rely on adaptation of FMs \cite{bommasani2022opportunities} for downstream tasks.
Generally, Large Language Models (LLMs) can achieve high performance on a downstream task relying only on their emergent abilities \cite{zhao2023survey}, without their weights being updated. Such adaption for a downstream task is called zero- or few-shot learning \cite{brown2020language} toward the text generation approaches (aka Generative AI or GenAI). In recent years, GenAI including models like Llama2 \cite{touvron2023llama}, Llama3 \footnote{https://github.com/meta-llama/llama3}, Mistral \cite{jiang2023mistral}, Mixtral \cite{jiang2024mixtral}. These models have gained the popularity in research areas and industries that revolutionized NLP applications by enabling high performance on various tasks through optimized prompting techniques rather than extensive fine-tuning. These models are pre-trained on vast and diverse datasets and can efficiently handle both simple and complex NLP tasks by simply refining prompts. This approach simplifies deployment and enhances adaptability, making instruct models particularly effective for tasks such as classification, text generation, summarization, and translation tasks.

In this work, we propose a fine-grainer detection of \textit{aspect category} into what we called \textit{entity} (i.e., meal) and \textit{aspect} (i.e., freshness) and focused on sentiment polarity revealed based on detected entity and aspect as a triplet output. This introduces a new ABSA subtask called ``Entity Aspect Sentiment Triplet Extraction (EASTE)'' task. We benchmark our experimentation settings on the ABSA subtask on the dataset published in a shared task; SemEval16 \cite{pontiki-etal-2016-semeval} which is a dataset based on restaurant reviews in English.
Our work also tackle these challenges following EASTE settings in different appraoches: 1) a token classification task like Named-entity recognition (NER) by proposing a unified-loss solution based on BERT model \cite{DBLP:journals/corr/abs-1810-04805} with full fine-tuning, 2) zero/few shot learning and fine-tuning text generation models and 3) model alignments via parameter efficient fine-tuning such as LoRA \cite{hu2021lora} and Prefix-tuning \cite{li2021prefixtuning} to observe the impacts and performances on complex tasks such as EASTE on instruct-tuned encoder-decoder LLMs.

We position these contributions on multiple fields.
\begin{itemize}
\item Creating an EASTE, a novel ABSA task that profoundly targets \textit{entity}, \textit{aspect} elements for comprehensive analysis of \textit{sentiment}.
\item Proposing a novel unified-loss approach to solve EASTE task via token classification that is suitable for multi-class classification.
\item Deeply exploring various techniques to adapt Language Models for a complex downstream task and achieving a State-of-the-Art (SoTA) results.
\end{itemize}

\section{Related Work}
\label{sec:related_work}
In this section, we highlight relevant research that shares the SoTA target sentiment analysis work. 
Regarding the ABSA subtasks which have been introduced as a task for SemEval14 \cite{pontiki-etal-2014-semeval} and continued to be present on both SemEval15 \cite{pontiki-etal-2015-semeval} and SemEval16 \cite{pontiki-etal-2016-semeval} competitions, most of complex SA tasks were introduced as compound ABSA subtasks \cite{zhang2022survey} which
aims to extract simultaneously multiple elements – pairs, triplets or quadruplets from a given sentence or a text. Among for our research relevant tasks such as  Aspect Category Sentiment Analysis (ACSA), Aspect Sentiment Triplet Extraction (ASTE), and Aspect Sentiment Quad Prediction (ASQP), we highlight Target-Aspect-Sentiment Detection (TASD).

\textbf{Target-Aspect-Sentiment Detection (TASD)}.
TASD aims to extract target, aspect, sentiment triplets. Target Aspect Sentiment Detection was introduced by \citet{Wan2020TargetAspectSentimentJD} and for a review sentence $S$ aims to extract all $(t, a, s)$ triplets: \[TASD(S) = \{(t_1, a_1, s_1), ..., (t_n, a_n, s_n)\}\] 
where $t_i$ stands for target and is a subsequence of $S$ if explicit and $NULL$ if implicit, $a_i \in \{a_1,...,a_n\}$ stands for aspect and $s_i \in \{positive, negative, neutral\}$ for sentiment polarity. For example, in the restaurant review sentence \textit{'The food arrived 20 minutes after I called, cold and soggy.'}, output of TASD should be \textit{\{(NULL, SERVICE\#GENERAL, negative), (food, FOOD\#QUALITY, negative)\}}

To our knowledge, \citet{brun-nikoulina-2018-aspect} and \citet{Wan2020TargetAspectSentimentJD} have addressed this problem using available parsers and domain-specific semantic lexicons and pre-trained BERT-based architecture respectively. In additional, the work of \citet{zhang-etal-2021-towards-generative} defines the TASD as sequence-to-sequence learning problem and solves it in a generative manner using encoder-decoder T5 architecture \cite{DBLP:journals/corr/abs-1910-10683}.

\section{Task Definition}

\citet{pontiki-etal-2015-semeval} define an entity \textit{e}, for instance in the domain of restaurants, as either the reviewed entity itself (restaurant), or another relevant entities directly related to it (e.g. food, service, ambience), while the aspect \textit{a} is a particular attribute (e.g., quality, price, general) of the entity in question. The sentiment polarity \textit{s} is defined as a polarity of the opinion(s) expressed towards the target. All three pieces of information are extracted from a predefined set of entities $e \in \{e_1,..., e_n\}$ and aspects $a \in \{a_1,..., a_n\}$ for every specific domain of interest, while sentiment polarity $s \in \{positive, negative, neutral\}$ remains the same per domain. \citet{Wan2020TargetAspectSentimentJD} combine what \citet{pontiki-etal-2015-semeval} defined as entity and aspect into one element as ``aspect'', taking into account only \textit{e} and \textit{a} combinations that appear in the analyzed datasets, which might not cover all true possible combinations.

Therefore, we follow the definitions by \citet{pontiki-etal-2015-semeval} and define new ABSA subtask which considers entity and aspect as two different elements. Hence, for a given sentence $S$ extracts all (e, a, s) triplets but taking into account that each of the triplets relates to the corresponding target:  
\[EASTE(S) = \{(t_1, e_1, a_1, s_1), ..., (t_n, e_n, a_n, s_n)\}\]
where $t_i$ stands for target and is a subsequence of $S$ if explicit and $NULL$ if implicit, $e_i \in \{e_1, ..., e_n\}$ stands for entity, $a_i \in \{a_1,...,a_n\}$ stands for aspect and $s_i \in \{positive, negative, netural\}$ for sentiment polarity. If we take the same restaurant sentence \textit{'The food arrived 20 minutes after I called, cold and soggy.'}, output of EASTE should be \textit{\{(NULL, SERVICE, GENERAL, negative), (food, FOOD, QUALITY, negative)\}}. Hence, we break the aspect into two elements as ``entity'' and ``aspect'' where each of them comes from different, independent dataset and is combined after its individual prediction in what \citet{Wan2020TargetAspectSentimentJD} define as aspect. This scenario of extraction creates one more level of ABSA task complexity.

\section{Dataset}
\label{sec:dataset}
Our work relies on the latest benchmark in this category which is SemEval16 \cite{pontiki-etal-2016-semeval}. We adopt the review sentences for restaurants domain. For each given sentence $(e, a, s)$ triplets are annotated where $e\in\{e_{1},e_{2},…e_{n}\}$ stands for entity, $a\in\{a_{1},a_{2},…a_{n}\}$ for aspect category, and $s\in\{positive,negative,neutral\}$ for sentiment. $e$, $a$ and $s$ are chosen from different predefined inventories. To each of the triplets, a \textit{target term} is attached towards which the opinion is expressed. If it is expressed explicitly, the \textit{target term} is a subset of the words of corresponding review sentence, and NULL if implicitly. The dataset contains 2000 review sentences for train set, and 676 review sentences for test set. 

\section{Methodology and Experimentation}
\label{sec:methodology}
In this section, we discuss our experimental settings to explore EASTE task while the token classification tasks were run locally using Apple CPU and MPS devices. For text generative tasks via LLMs like Llama2, Llama3, and Mixtral models, we use API calls for model inference hosted via IBM watsonx.ai\footnote{https://www.ibm.com/products/watsonx-ai}.

\subsection{Classification Approach}

\begin{figure}[h!]
  \includegraphics[scale=0.29]{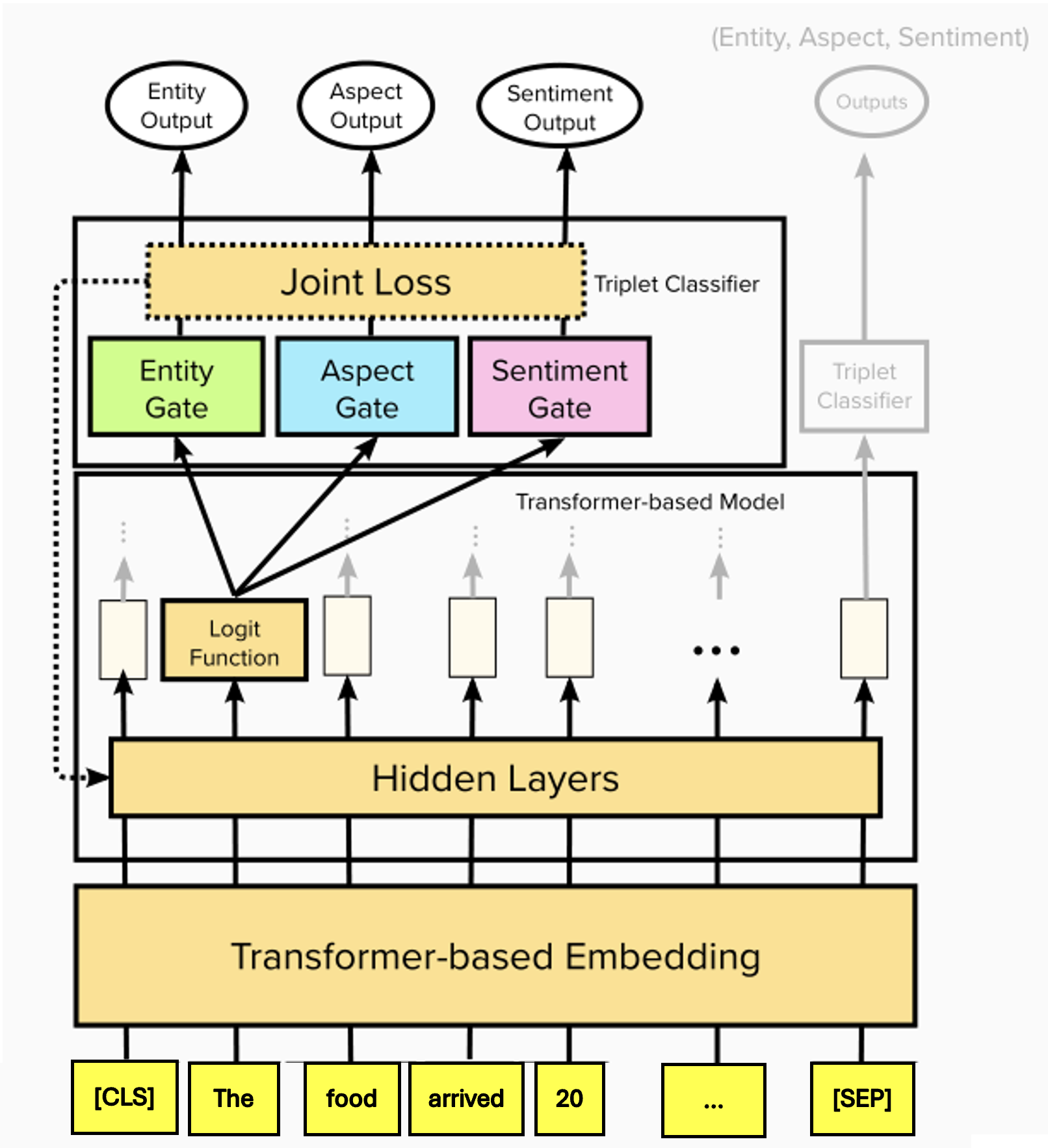}
  \caption{Unified-loss architecture based on BERT token classification where each loss function of $entity, aspect, sentiment$ are weighted thoroughly.}
  \label{fig:unified_loss}
\end{figure}

We solve EASTE task using a proposed unified-loss approach toward token classification task which is applied on a sentence $S$ using BERT architecture.
Figure \ref{fig:unified_loss} demonstrates our implementation of the unified-loss approach based on token classification where a triple classifier is introduced after a loss function per gate.\footnote{Unified-loss code snippet is shared via GitHub link: https://github.com/vvorakit/Entity-Aspect-Sentiment-Triplet-Extraction} Each token in $S$ represents $\{t_1, t_2, ..., t_n\}$, every $t$ provides single or multiple $(entity, aspect, sentiment)$ triplets where $t=\{(e_1, a_1, s_1),...,(e_n, a_n, s_n)\}$.
We  modify the last linear layer of BERT-based-uncased architecture and consequently adapt model's loss function to obtain three classification results where the losses $l$ are computed as an average loss $l(joint)$. The final loss is calculated as a mean of losses via the additional of loss logits then divided by number of output gate of each $entity, aspect, sentiment$ as following:
\[l(joint) = \frac{l(entity)+l(aspect)+l(sentiment)}{Number\ of\ output\ gates}\]
where $l(entity), l(aspect), l(sentiment)$ are individual cross entropy losses for entity, aspect and sentiment obtained per token.

\textbf{Setting 1: Entity-Aspect Sentiment Triplet Extraction}
For the unified-loss approach explained aforementioend, we adopted BERT base uncased (110M) for token classification task and fine-tuned it for 50 epochs. The batch size we used for this approach was 1, combined with stochastic gradient descent (SGD) optimization algorithm with learning rate of 1e-3. Batch size, optimization algorithm used are not very common in practice nowadays, but in this combination they proved to obtain the best results. The measure of quality are precision, recall and F1 score.. The best results on token level which are presented in the Table \ref{table:unified-loss} are obtained in the epoch 48.

\subsection{Text Generation Approach}
In the Text Generation Task framework, models generate an output in dictionary format from an input prompt\footnote{All prompts from GenAI models used in EASTE task can be found in Appendix.} which consist of input sentence $S$ and instruction $Inst$. GenAI models for EASTE task aim to capture information on specific triplets containing entities, aspects, and sentiment polarity. For each input sentence $S$, the model generates $\{t_1, t_2, ..., t_n\} $ tokens where $ n $ varies for different input $t$ as output, then we encode them into text format. Using zero and few-shot learning approaches and fine-tuning for a downstream task, the model is trained so that the combined outputs $ \{t_1, t_2, ..., t_n\} $ contain information $ \{(e_1,a_1,s_1),...,(e_n,a_n,s_n)\} $ triplets only. Unlike Token Classification Tasks, where tokens are assigned to classes, text generative models ensure that the extracted triplets correspond accurately to target terms within review sentences. Evaluating zero and few-shot learning approaches, we assess the capability of LLMs to solve complex tasks like EASTE solely based on pre-training. Zero-shot learning involves providing the model with instructions and review sentences, expecting output to contain relevant triplets referencing target terms $LM(S, Inst) = \{(t_1, e_1, a_1, s_1), ..., (t_n, e_n, a_n, s_n)\} $. Few-shot learning extends this by including input-output examples, maintaining a constant set of examples while varying input sentences $ LM(S, (Inst, \{e_1,...,e_n\})) = \{(t_1, e_1, a_1, s_1), ..., (t_n, e_n, a_n, s_n)\} $.

\textbf{Setting 2: Text Generation with Zero/Few-shot Technique}
In the zero/few-shot setting, we conducted numerous experiments searching for the instruction that provides the best results. We experimented with the instructions that provide from 0 (zero-shot) gradually up to 40 (few-shot) review sentence examples from a benchmark dataset to observe learning capabilities of the model. 

\textbf{Setting 3: Text Generation with Fine-tuning Technique}

In the fine-tuning setting, we conducted numerous experiments searching for the overall best practice in terms of the model size, memory and time consumption, type of the instruction, as well as the performance on the task. The approaches that we use for fine-tunning are full model fine-tuning which updates all model's trainable parameters via back-propagation, Prefix-tuning \cite{DBLP:journals/corr/abs-2101-00190} and LoRA \cite{hu2021lora} that update only the small percentage of them (form 0.1\% to 0.3\% of number of trainable parameters). 

For full model fine tuning Flan-T5-XL (3B) trained for 6 epochs with batch size of 4 and AdamW optimizer with learning rate of 2e-4 performed the best. It is important to note that we did not apply the full model fine-tuning on models larger than 3B parameters due to the memory constraints. Furthermore, in the Prefix-tuning setting, our best performing model was Flan-UL2 (20B) which was trained for 50 epochs, using as well the batch size of 4 combined with AdamW optimizer with 2e-4 learning rate, and setting the number of virtual tokens to 30. Finally, the best performing LoRA tuned model was Flan-T5-XXL (11B) which was trained for 10 epochs as well with the batch size of 4 and AdamW optimizer with learning rate of 2e-4. In LoRA setting, we approximate only query and value update matrices with rank 8, and set alpha to 16. All other non-discussed training parameters are set to default values. The best performing results are represented in Table \ref{table:full_comparison}

\section{Results Analysis}

\begin{table}[t]
\centering
\begin{tabularx}{\columnwidth}{|>{\centering\arraybackslash}l|>{\centering\arraybackslash}X|>{\centering\arraybackslash}X|>{\centering\arraybackslash}X|}
\hline
\textbf{EASTE Task} & \textbf{P} & \textbf{R} & \textbf{F1} \\ 
\hline
\textbf{Unified-loss-BERT-uncased} & 57.41 & 54.25 & 55.78 \\
\hline
\textbf{Zero-shot prompting} & & & \\
Llama2-13b-chat & 45.72 & 99.13 & 43.50 \\
Llama3-8b-instruct & 45.79 & 97.36 & 62.28 \\
Mixtral-8x7B-Instruct-v0.1 & 46.32 & 99.47 & 63.21 \\
\hline
\textbf{One-shot prompting} & & & \\
Llama2-13b-chat & 37.50 & 99.35 & 54.45 \\
Llama3-8b-instruct & 57.17 & 99.55 & 70.16 \\
Mixtral-8x7B-Instruct-v0.1 & 54.03 & \textbf{100} & 70.16 \\
\hline
\textbf{Few-shot prompting} & & & \\
Llama2-13b-chat & 45.72 & \textbf{100} & 62.75 \\
Llama3-8b-instruct & 54.28 & \textbf{100} & 70.36 \\
Mixtral-8x7B-Instruct-v0.1 & 55.01 & \textbf{100} & 70.98 \\
\hline
\textbf{Full fine-tuning} & & & \\
Flan-T5-XL & \textbf{76.87} & 75.90 & \textbf{76.38} \\
\hline
\textbf{Prefix fine-tuning} & & & \\
Prefix tuning Flan-UL2 & 68.04 & 62.74 & 65.28 \\
\hline
\textbf{LoRa fine-tuning} & & & \\
LoRA-Flan-T5-XXL & 69.87 & 63.47 & 66.52 \\
\hline
\end{tabularx}
\caption{Reporting only best results for EASTE task per approach across multiple techniques on SemEval2016 restaurant test set.}
\label{table:full_comparison}
\end{table}

\begin{table}[t]
\centering
\begin{tabularx}{\columnwidth}{|>{\centering\arraybackslash}l|>{\centering\arraybackslash}X|
>{\centering\arraybackslash}X|
>{\centering\arraybackslash}X|}
\hline
\textbf{TASD Task} & \textbf{P} & \textbf{R} & \textbf{F1} \\
\hline
\textbf{Entity\#Aspect} &  69.32 & 58.82& 63.64 \\
\textbf{Sentiment} &  66.96 & 68.80 & 87.87 \\
\hline
\end{tabularx}
\caption{Results on token classification task based on TASD settings and chained evaluation of 2 outputs as Entity\#Aspect and sentiment via unified-loss approach using BERT architecture on the SemEval2016 restaurants test set.}
\label{table:TASD-unified-loss}
\end{table}

\begin{table}[t]
\centering
\begin{tabularx}{\columnwidth}{|>{\centering\arraybackslash}l|>{\centering\arraybackslash}X|>{\centering\arraybackslash}X|>{\centering\arraybackslash}X|}
\hline
\textbf{ABSA Task} & \textbf{P} & \textbf{R} & \textbf{F1} \\
 \hline
\textbf{Entity} &  75.90 & 69.19& 72.39 \\
\textbf{Aspect} &  53.10  & 52.74  & 52.92  \\
\textbf{Sentiment} &  62.51 & 55.33 & 58.70 \\
\hline
\end{tabularx}
\caption{Results on token classification task based on ABSA settings for individual evaluation of entity, aspect, and sentiment via unified-loss approach using BERT architecture on the SemEval2016 restaurants test set.}
\label{table:unified-loss}
\end{table}

The main indicator of model performance in our experimentation is F1 score. For each of methodology settings, we defined a way to calculate F1 score in order to get comparable results. In general, the (e, a, s) triplets are considered predicted correctly if and only if the prediction is attached to the correct corresponding \textit{target term}. 

\textbf{Setting 1:} As the \textit{target term} is composed of either one or multiple words if the opinion expressed explicitly towards entity, we consider prediction correct only if 50\% or more tokens of the \textit{target term} are predicted correctly. On the other hand, if the opinion is expressed implicitly we expect model to attach (e, a, s) triplet to \textit{cls} token. 

\textbf{Settings 2 and 3:} We consider prediction correct only if (e, a, s) triplet is generated attached to correct corresponding \textit{target term} if opinion expressed explicitly towards an entity, otherwise we expect a \textit{target term} to be predicted as \textit{'NULL'}. 

Overall, our best performing model is fully fine-tuned Flan-T5-XL achieving F1 score of 76.38 based on proposed EASTE experimental settings. However, a few-shot prompted on recently released LLMs such as Llama2, Llama3, and Mixtral obtain outstanding results on recall, but considerably lower on precision which impact the final F1-score. Additionally, Mixtral performs as second ranked, up close by Llama3 with 70.98 and 70.36 F1 scores respectively. On the other hand, prefix-tuning yielded competitive results where Flan-UL2 emerged as a top-performing model, achieving a F1 score of 65.28. Our exploration of LoRA tuning did not provide the big gap over prefix-tuning, with the best scoring model being Flan-T5-XXL with the F1 score of 69.87. However, the full-tuned encoder models for token classification using unified-loss approach tends to provide reasonably good results due to joint layers giving the high complexity of the prediction. 
On the other hand, in Table \ref{table:TASD-unified-loss} when we cast the unified-loss approach to evaluate in TASD settings where entity and aspect are merged as a single element, it shows that there is a significant improvement up to almost 10\% in the classification report.
Ultimately, if we split an evaluation category into individual inspection of Entity, Aspect, Sentiment like in Table \ref{table:unified-loss}, we can observe that the unified-loss approach using BERT is well performed on pre-defined entity detection with F1 score of 72.39 in token classification settings where aspect and sentiment remains a challenging detection process.

\section{Conclusion}
This research introduces EASTE, a novel and complex task for detecting ABSA settings through various NLP techniques. Our results underscore the critical importance of selecting appropriate fine-tuning techniques and prompting strategies tailored to the size and type of LLMs. By employing diverse approaches such as token classification, text generation, and fine-tuning alignments, we assessed model performance across different architectures and sizes. Our findings demonstrate the efficacy of the proposed unified-loss approach in token classification, particularly for less complex tasks like TASD. Additionally, the potential of text generative models combined with advanced prompting strategies is evident. We also highlight the necessity of choosing suitable fine-tuning techniques and model architectures based on task complexity and available resources. This research offers significant contributions to the field of sentiment analysis, providing deeper insights into sophisticated NLP techniques and models for complex sentiment analysis tasks. Our work advances the understanding of deep content analysis for sentiment detection and sets a foundation for future explorations in the realm of natural language processing.

\section*{Acknowledgements}
We would like to thank IBM Client Engineering EMEA for providing the GPU infrastructure that made our research possible. Their support enabled us to carry out our experiments efficiently and effectively. The computational resources provided by IBM were instrumental in advancing our work, especially in model alignments and the diverse fine-tuning approaches utilized in this study.

\nocite{langley00}

\bibliography{icml2024}
\bibliographystyle{icml2024}

\newpage
\appendix
\onecolumn
\section{Appendix.}

Prompts used in Entity Aspect Sentiment Triplet Extraction (EASTE) task are listed below.

\begin{enumerate}
\item Flan-T5, Tk-Instruct and Flan-UL2 models: \\
\begin{lstlisting}
Definition: In this task you are given a review sentence and your task is to extract the triplet of information 'entity':'aspect':'sentiment' for each 'term' (implicit or explicit) the opinion is expressed towards in the given review sentence. The final output should be in shape 'term':'entity':'aspect':'sentiment'. Every implicit 'term' should be classified as 'NULL'.
  
Example 1- 
Input: great food, great wine list, great service in a great neighborhood...
Output: food:food:quality:positive, wine list:drinks:style_options:positive, service:service:general:positive, neighborhood:location:general:positive
  
Example 2-
Input: Rather than preparing vegetarian dish, the chef presented me with a plate of steamed vegetables (minus sauce, seasoning, or any form or aesthetic presentation).
Output: vegetarian dish:food:quality:negative, vegetarian dish:food:style_options:negative, chef:service:general:negative
  
Example 3- 
Input: The chicken lollipop is my favorite, most of the dishes (I have to agree with a previous reviewer) are quite oily and very spicy, especially the Chilli Chicken.
Output: chicken lollipop:food:quality:positive, dishes:food:quality:negative, Chilli Chicken:food:quality:negative
  
Example 4- 
Input: Also, they do not take credit card so come with cash!
Output: NULL:restaurant:miscellaneous:neutral
  
Example 5-
Input: The appetizers we ordered were served quickly - an order of fried oysters and clams were delicious but a tiny portion (maybe 3 of each). 
Output: fried oysters and clams:food:quality:positive, fried oysters and clams:food:style_options:negative, NULL:service:general:positive
  
Example 6-
Input: The service was spectacular as the waiter knew everything about the menu and his recommendations were amazing!
Output: service:service:general:positive, waiter:service:general:positive
  
Example 7-
Input: I book a gorgeous white organza tent which included a four course prix fix menu which we enjoyed a lot.
Output: white organza tent:ambience:general:positive, four course prix fix menu:food:quality:positive
  
Example 8- 
Input: The place is beautiful!
Output: place:ambience:general:positive
  
Example 9- 
Input: MY husbands birthday and my sons was not as it was intended... and we drove two hours to spend too much money to be treated terribly!
Output: NULL:restaurant:general:negative, NULL:restaurant:prices:negative, NULL:service:general:negative
  
Now complete the following example- 
Input: {sentence}
Output:
\end{lstlisting}

\item Llama2-13b-chat model: \\
\begin{lstlisting}
<s>[INST] <<SYS>> You are a cautious assistant.You follow  strictly the prompt. You carefully follow instructions. You are helpful and harmless and you follow ethical guidelines and promote positive behavior. If you don't know the answer to a question, please don't share false information. <</SYS>>

A triplet is a set of three elements: an entity (E), an attribute (A), and a sentiment (S). Your task is to generate only one (entity, attribute, sentiment) found in the given sentence. In each sentence , you must find exactly one triplet.

The entity must be chosen from the list  ['FOOD', 'RESTAURANT', 'SERVICE', 'AMBIENCE', 'DRINKS', 'LOCATION'].

The attribute must be chosen from the list ['QUALITY', 'STYLE\_OPTIONS', 'GENERAL', 'PRICES', 'MISCELLANEOUS'].

The sentiment must be chosen from ['positive', 'negative', 'neutral'].

Don't generate any text other than the JSON dictionnary.
JSON Format for triplet prediction:
{
  "triplet": 
    {
      "entity": "ENTITY_TYPE",
      "attribute": "ATTRIBUTE_TYPE",
      "sentiment": "SENTIMENT_TYPE"
    }
}
Replace ENTITY_TYPE with exactly one of the predefined entity types ( ['FOOD', 'RESTAURANT', 'SERVICE', 'AMBIENCE', 'DRINKS', 'LOCATION']), ATTRIBUTE_TYPE with exactly one of the attribute types(['QUALITY', 'STYLE_OPTIONS', 'GENERAL', 'PRICES', 'MISCELLANEOUS']), and SENTIMENT_TYPE with exactly  one of the sentiment types (["positive", "negative",  "neutral"]).
Respect the given format.
Sentence:
\end{lstlisting}

\item Llama3-8b-instruct: \\
\begin{lstlisting}
<|begin_of_text|> <|start_header_id|> system <|end_header_id|> You are a cautious assistant. You follow strictly the prompt. You carefully follow instructions. You are helpful and harmless and you follow ethical guidelines and promote positive behavior. If you don't know the answer to a question, please don't share false information. <|eot_id|>

<|begin_of_text|> <|start_header_id|> user <|end_header_id|> 
A triplet is a set of three elements: an entity (E), an attribute (A), and a sentiment (S). Your task is to generate only one (entity, attribute, sentiment) found in the given sentence. In each sentence , you must find exactly one triplet.

The entity must be chosen from the list  ['FOOD', 'RESTAURANT', 'SERVICE', 'AMBIENCE', 'DRINKS', 'LOCATION'].

The attribute must be chosen from the list ['QUALITY', 'STYLE_OPTIONS', 'GENERAL', 'PRICES', 'MISCELLANEOUS'].

The sentiment must be chosen from ['positive', 'negative', 'neutral'].

Don't generate any text other than the JSON dictionary.
JSON Format for triplet prediction:
{
  "triplet": 
    {
      "entity": "ENTITY_TYPE",
      "attribute": "ATTRIBUTE_TYPE",
      "sentiment": "SENTIMENT_TYPE"
    }
}
Replace ENTITY_TYPE with one of the predefined entity types, ASPECT_TYPE with one of the attribute types, and SENTIMENT_TYPE with one of the sentiment types.
Respect the given format.
Sentence: <|eot_id|>
\end{lstlisting}

\item Mixtral-8x7B-Instruct-v0.1 model: \\
\begin{lstlisting}
<|system|> You are a cautious assistant. You carefully follow instructions. You are helpful and harmless and you follow ethical guidelines and promote positive behavior. If a question does not make any sense, or is not factually coherent, explain why instead of answering something not correct. If you don't know the answer to a question, please don't share false information.

<|user|> A triplet is a set of three elements: an entity (E), an attribute (A), and a sentiment (S). Your task is to generate only one triplet (entity, attribute, sentiment) from the given sentence.

The entity must be chosen from the predefined entity types ['FOOD', 'RESTAURANT', 'SERVICE', 'AMBIENCE', 'DRINKS', 'LOCATION'].
The attribute must be chosen from the list ['QUALITY', 'STYLE_OPTIONS', 'GENERAL', 'PRICES', 'MISCELLANEOUS'].
The sentiment must be chosen from ['positive', 'negative', 'neutral'].

Your response must be in JSON format, correctly written and complete. Don't forget the braces. Don't add any comments at all. Only the triplet is required.
Format for triplet prediction:
{
  "triplet": 
    {
      "entity": "ENTITY_TYPE",
      "attribute": "ATTRIBUTE_TYPE",
      "sentiment": "SENTIMENT_TYPE"
    }
  
}
Replace ENTITY_TYPE with one of the predefined entity types, ATTRIBUTE_TYPE with one of the attribute types, and SENTIMENT_TYPE with either "positive", "negative", or "neutral".
Sentence: 
\end{lstlisting}

\end{enumerate}

\end{document}